\documentclass[conference]{IEEEtran}
\IEEEoverridecommandlockouts

\usepackage{cite}
\usepackage{amsmath,amssymb,amsfonts}
\usepackage{algorithmic}
\usepackage{graphicx}
\usepackage{textcomp}
\usepackage{xcolor}
\usepackage{booktabs} 
\usepackage{algorithm} 
\usepackage{hyperref}
\usepackage{pgfplots}
\pgfplotsset{compat=1.17} % or newer
\usetikzlibrary{patterns}
\def\BibTeX{{\rm B\kern-.05em{\sc i\kern-.025em b}\kern-.08em
    T\kern-.1667em\lower.7ex\hbox{E}\kern-.125emX}}

\begin{document}

\title{Post-Pruning Accuracy Recovery via Data-Free Knowledge Distillation}

\author{\IEEEauthorblockN{Chinmay Tripurwar, Utkarsh Maurya, and Dishant}
\IEEEauthorblockA{\textit{Department of Electrical Engineering} \\
\textit{Indian Institute of Technology Bombay}\\
Mumbai, India \\
\{22B3902, 22B3910, 22B984\}@iitb.ac.in}
}

\maketitle

\begin{abstract}
Model pruning is a widely adopted technique to reduce the computational complexity and memory footprint of Deep Neural Networks (DNNs). However, global unstructured pruning often leads to significant degradation in accuracy, typically necessitating fine-tuning on the original training dataset to recover performance. In privacy-sensitive domains such as healthcare or finance, access to the original training data is often restricted post-deployment due to regulations (e.g., GDPR, HIPAA). This paper proposes a Data-Free Knowledge Distillation framework to bridge the gap between model compression and data privacy. We utilize DeepInversion to synthesize privacy-preserving ``dream'' images from the pre-trained teacher model by inverting Batch Normalization (BN) statistics. These synthetic images serve as a transfer set to distill knowledge from the original teacher to the pruned student network. Experimental results on CIFAR-10 across various architectures (ResNet, MobileNet, VGG) demonstrate that our method significantly recovers accuracy lost during pruning without accessing a single real data point.
The code is available at: \href{https://github.com/chinoscode1708/DF-Prune-Recover}{https://github.com/chinoscode1708/DF-Prune-Recover}
\end{abstract}

\begin{IEEEkeywords}
Model Compression, Neural Network Pruning, Knowledge Distillation, Data-Free Learning, DeepInversion.
\end{IEEEkeywords}

\section{Introduction}
Deep Convolutional Neural Networks (CNNs) have achieved state-of-the-art performance in computer vision tasks. However, their deployment on edge devices (e.g., IoT sensors, mobile phones) is often hindered by high computational and memory costs. Network pruning addresses this by removing redundant weights, but aggressive pruning (e.g., 50\% sparsity) often degrades model inference capabilities, rendering the model unusable without retraining.

Standard recovery methods involve fine-tuning the pruned model using the original dataset. However, a growing number of real-world scenarios prohibit access to training data after the model is trained. This restriction arises due to:
\begin{enumerate}
    \item \textbf{Privacy Concerns:} In medical imaging, patient data cannot leave secure servers due to HIPAA/GDPR compliance.
    \item \textbf{Data Size:} The original training set (e.g., JFT-300M) may be too large to distribute to edge engineers optimizing the model.
    \item \textbf{Intellectual Property:} The dataset itself may be proprietary, while the model weights are shared.
\end{enumerate}

To address this dilemma, we introduce a pipeline that recovers the accuracy of a pruned model using only the model weights itself. Our approach leverages the pre-trained model as a "Teacher" to generate synthetic data that matches the statistical distribution of the original training set. This is achieved via \textit{DeepInversion}, which optimizes random noise to match the Batch Normalization (BN) statistics stored in the teacher. We then employ Knowledge Distillation (KD) to transfer the teacher's robust representations to the pruned "Student," effectively recovering accuracy in a completely data-free setting.

\section{Related Work}

\subsection{Network Pruning}
Pruning reduces model size by setting specific weights to zero. Structured pruning removes entire channels or filters, allowing for immediate speedups on hardware but often causing larger accuracy drops at high compression rates. Unstructured pruning removes individual weights based on magnitude. In this work, we focus on global unstructured pruning using the L1-norm criterion, which generally yields higher sparsity but requires robust recovery mechanisms.
\begin{figure}[htbp]
    \centering
    % REPLACE WITH ACTUAL DIAGRAM
    \includegraphics[width=0.95\linewidth]{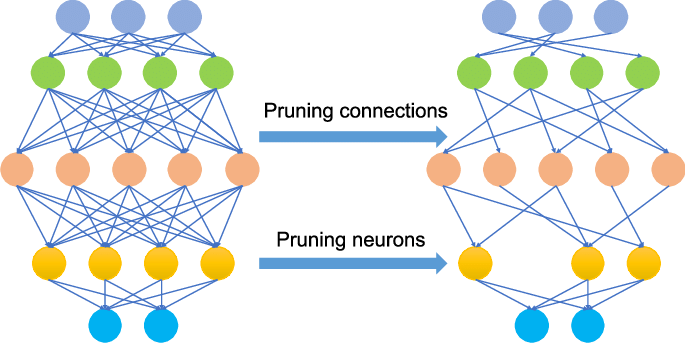}
    \caption{Visualization of Neural Network Pruning. The left diagram represents a dense network, while the right represents the sparse network after global unstructured pruning.}
    \label{fig:pruning_viz}
\end{figure}

\subsection{Data-Free Knowledge Distillation}
Knowledge Distillation (KD) transfers information from a large Teacher to a smaller Student. Traditional KD requires a proxy dataset. Recent "Data-Free" approaches generate this proxy dataset synthetically. Methods like DeepInversion \cite{b2} optimize input noise to minimize the distance between the feature statistics of the input and the running statistics stored in the network's BatchNorm layers. This allows the synthesis of images that are statistically similar to the training data without containing actual private information.

\begin{figure}[htbp]
    \centering
    % REPLACE WITH ACTUAL DIAGRAM
    \includegraphics[width=0.95\linewidth]{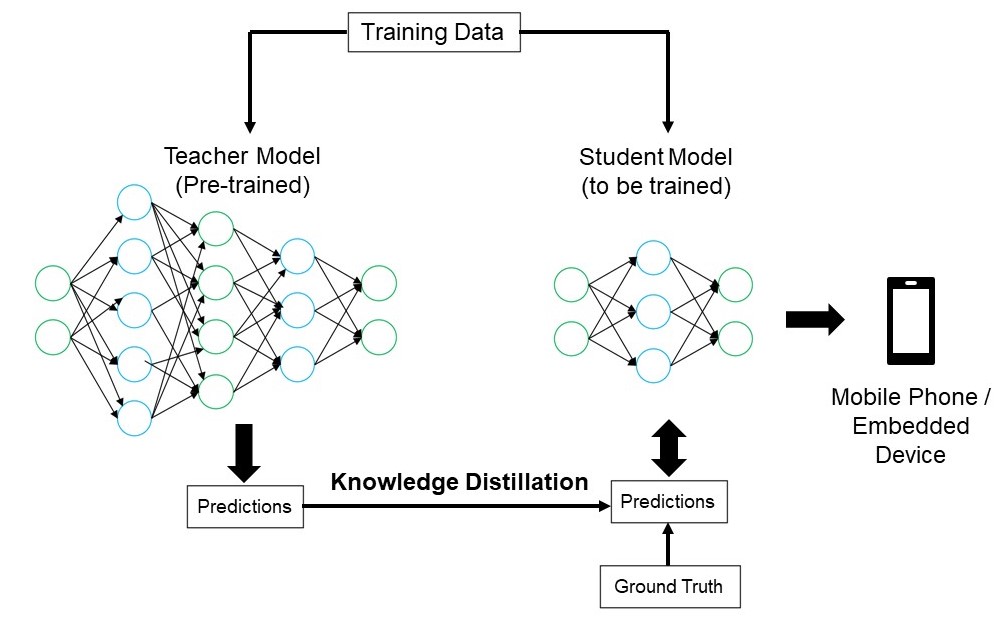}
    \caption{Overview of the Knowledge Distillation process where the Teacher guides the Student network.}
    \label{fig:kd_viz}
\end{figure}

\section{Methodology}

Our framework consists of three sequential stages: (1) Teacher Preparation and Pruning, (2) DeepInversion-based Data Synthesis, and (3) Recovery via Distillation.

\subsection{Pruning Strategy}
Let $T$ be the pre-trained teacher model parameterized by weights $W_T$. We derive a sparse student model $S$ by applying global unstructured pruning. We calculate the L1-norm of all weights in convolutional and linear layers and mask the bottom $p\%$ (in our experiments, $p=50$).
\begin{equation}
    W_S = m \odot W_T
\end{equation}
where $m \in \{0, 1\}$ is a binary mask. The mask is computed globally across all applicable layers, allowing the algorithm to prune layers with less importance more aggressively while preserving sensitive layers. At this stage, without fine-tuning, the accuracy of $S$ typically drops by 40-80\% depending on the architecture redundancy.

\subsection{Synthetic Data Generation (DeepInversion)}
To recover $S$ without real data, we generate a synthetic dataset $\mathcal{D}_{syn}$. We initialize a batch of random noise inputs $\hat{x} \sim \mathcal{N}(0, I)$. We optimize $\hat{x}$ via gradient descent (holding the Teacher weights fixed) to minimize a composite loss function.

The optimization objective is:
\begin{equation}
    \mathcal{L}_{syn} = \mathcal{L}_{CE} + \lambda_{BN}\mathcal{L}_{BN} + \lambda_{TV}\mathcal{L}_{TV}
\end{equation}

\begin{figure*}[t]
    \centering
    % REPLACE WITH ACTUAL DIAGRAM
    \includegraphics[width=1.0\textwidth]{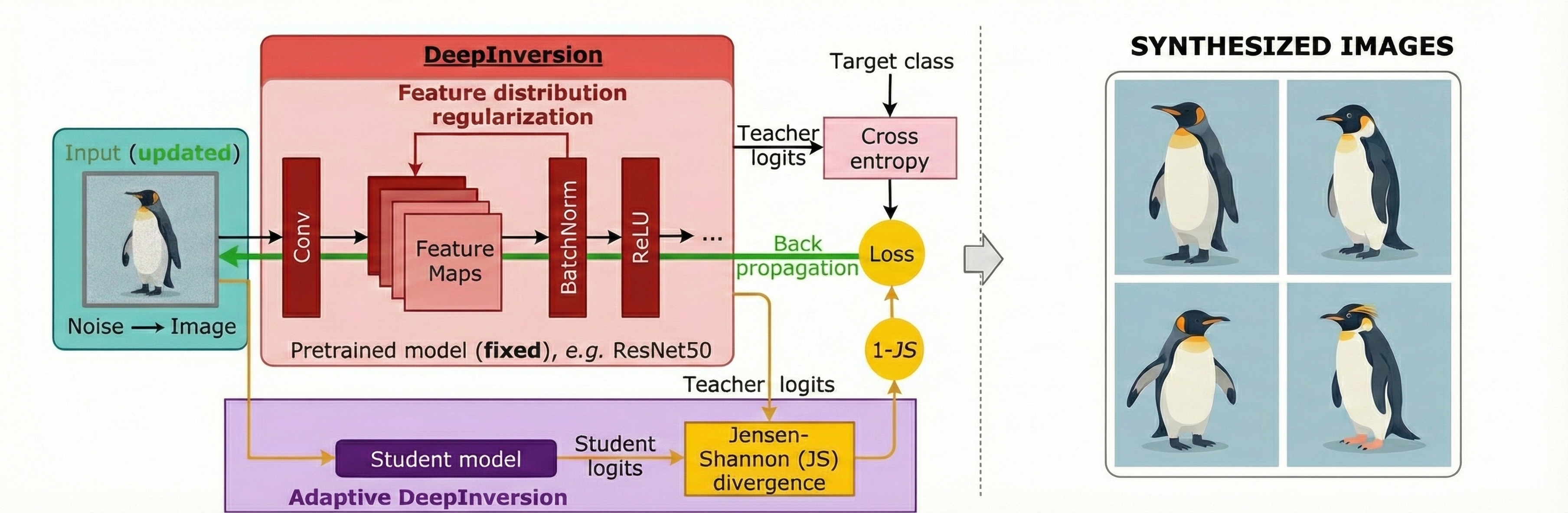} 
    \caption{DeepInversion Pipeline}
    \label{fig:kd_viz}
\end{figure*}

\subsubsection{Cross-Entropy Loss ($\mathcal{L}_{CE}$)}
We minimize the entropy of the teacher's predictions to encourage class-consistent images. A random noise image usually results in a uniform probability distribution output. By forcing the teacher to be confident in its prediction (even if the class is arbitrary), the generated image develops distinctive features of that class.
\begin{equation}
    \mathcal{L}_{CE} = - \sum_{k} p_T(k|\hat{x}) \log p_T(k|\hat{x})
\end{equation}

\subsubsection{BatchNorm Feature Loss ($\mathcal{L}_{BN}$)}
This is the core component of DeepInversion. The teacher model contains running means ($\mu_{run}$) and variances ($\sigma^2_{run}$) in every Batch Normalization layer, which represent the statistics of the original CIFAR-10 data. We enforce that the statistics of the generated batch match these stored values:
\begin{equation}
    \mathcal{L}_{BN} = \sum_{l} ||\mu_l(\hat{x}) - \mu_{l, run}||_2 + ||\sigma^2_l(\hat{x}) - \sigma^2_{l, run}||_2
\end{equation}
By satisfying this constraint across all layers (deep and shallow), the synthetic image $\hat{x}$ is forced to adopt the texture and structural properties of real data.

\subsubsection{Total Variation ($\mathcal{L}_{TV}$)}
To encourage spatial smoothness and natural-looking images, we apply Total Variation regularization. This penalizes high-frequency noise between adjacent pixels:
\begin{equation}
    \mathcal{L}_{TV} = \sum_{i,j} \left( |x_{i+1,j} - x_{i,j}| + |x_{i,j+1} - x_{i,j}| \right)
\end{equation}
Additionally, we employ random image jittering (shifting pixels horizontally/vertically) during the optimization steps to improve the robustness of the generated features.

\subsection{Recovery via Distillation}
Once $\mathcal{D}_{syn}$ is generated, we freeze the Teacher $T$ and train the pruned Student $S$. Since the synthetic labels are soft probabilities from the Teacher, we use KL Divergence loss:
\begin{equation}
    \mathcal{L}_{KD} = \alpha T^2 \cdot \mathcal{KLD}\left(\sigma\left(\frac{z_S}{T}\right), \sigma\left(\frac{z_T}{T}\right)\right)
\end{equation}
where $T=3.0$ is the temperature parameter used to soften the probability distributions, and $\sigma$ is the softmax function.
\begin{figure}[H]
    \centering
    % REPLACE WITH ACTUAL DIAGRAM
    \includegraphics[width=1.0\linewidth]{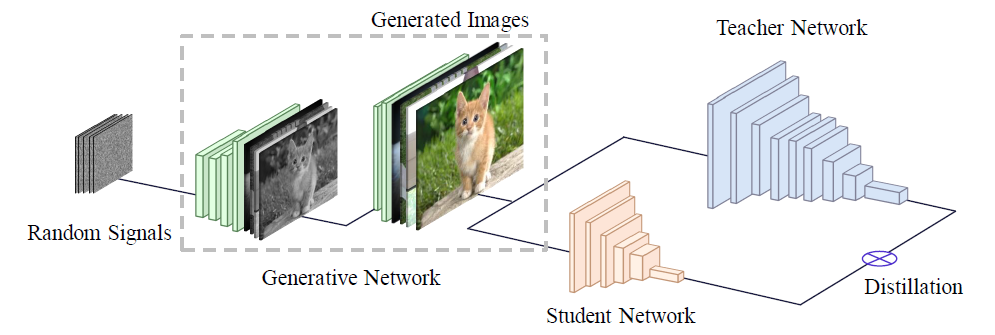}
    \caption{The proposed pipeline: (A) Teacher Inversion to create synthetic data, followed by (B) Student Recovery via Distillation.}
    \label{fig:pipeline}
\end{figure}
\textbf{Crucial Implementation Detail:} During the recovery phase, we set the BatchNorm layers of the student to \texttt{eval} mode. This freezes the student's running statistics. If we allowed the student to update its BN stats on synthetic data, it might drift away from the true data distribution, as synthetic data is an approximation. The student only updates its weights (Conv/Linear) to mimic the teacher's behavior.

\begin{algorithm}[h]
\caption{Data-Free Recovery Pipeline}
\begin{algorithmic}[1]
\REQUIRE Pre-trained Teacher $T$, Pruning Amount $p$
\STATE \textbf{Step 1:} Prune $T$ to create Student $S$ with $p\%$ sparsity.
\STATE \textbf{Step 2:} Generate $\mathcal{D}_{syn}$
\FOR{$i=1$ to $NumImages$}
    \STATE Init noise $\hat{x}$
    \STATE Optimize $\hat{x}$ using $\nabla_{\hat{x}}(\mathcal{L}_{CE} + \mathcal{L}_{BN} + \mathcal{L}_{TV})$
    \STATE Apply Random Jitter to $\hat{x}$
    \STATE Store $\hat{x}$ in $\mathcal{D}_{syn}$
\ENDFOR
\STATE \textbf{Step 3:} Distill Knowledge
\FOR{batch $(\hat{x})$ in $\mathcal{D}_{syn}$}
    \STATE Compute soft targets $y_T = T(\hat{x})$
    \STATE Train $S$ to minimize $\mathcal{L}_{KD}(S(\hat{x}), y_T)$
    \STATE \textit{Note: Keep BN layers of $S$ frozen.}
\ENDFOR
\STATE \textbf{Return} Recovered Student $S$
\end{algorithmic}
\end{algorithm}

\section{Experiments}

\subsection{Experimental Setup}
We implemented the pipeline using PyTorch on an NVIDIA GeForce GPU. The setup ensures reproducibility through fixed seeds.
\begin{itemize}
    \item \textbf{Dataset:} CIFAR-10 (used for Teacher pre-training only). The dataset contains 50,000 training images and 10,000 test images across 10 classes.
    \item \textbf{Architectures:} We evaluated on the resnet family of neural networks , ResNet (18, 34, 50).
    \item \textbf{Pruning:} Global L1 Unstructured pruning at 75\% sparsity.
    \item \textbf{Synthesis Hyperparameters:} We generated 1,024 synthetic images per model. The optimization ran for 200 iterations with a Learning Rate (LR) of 0.05. The weighting factors were $\lambda_{BN}=10.0$ and $\lambda_{TV}=1e^{-5}$.
    \item \textbf{Recovery Hyperparameters:} The pruned models were fine-tuned on the synthetic dataset for 15 epochs using Stochastic Gradient Descent (SGD) with a learning rate of 0.001, momentum of 0.9, and batch size of 32.
\end{itemize}

\subsection{Results}

\begin{figure}[htbp]
    \centering
    \includegraphics[width=1\columnwidth]{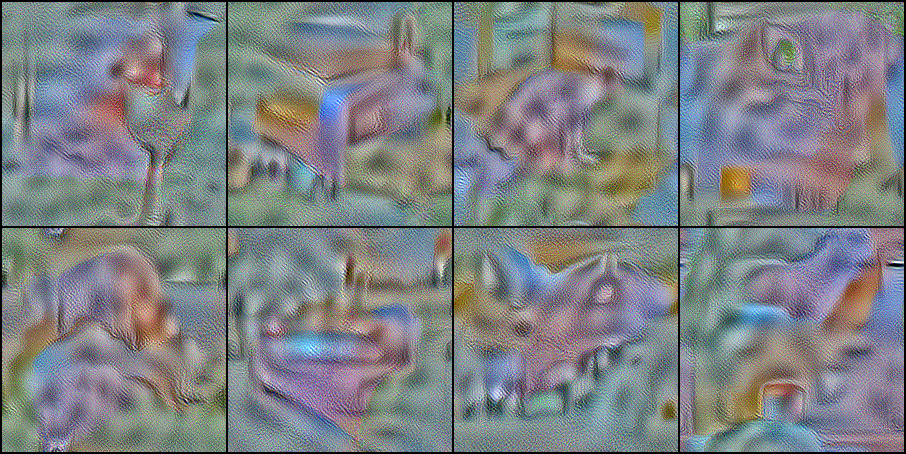}
    % \placeholderImage{Insert Figure: Grid of synthetic 'Dream' images generated by DeepInversion}{4cm}
    \caption{Visualization of synthetic images recovered from ResNet50 BN statistics. While not photorealistic, these images capture the dominant frequency and color statistics required for the student to learn decision boundaries.}
    \label{fig:dreams}
\end{figure}

We evaluated the classification accuracy on the CIFAR-10 test set at three stages: (1) Original Teacher accuracy, (2) Pruned Model (Zero-Shot) accuracy immediately after masking weights, and (3) Recovered Model accuracy after distillation.

Table \ref{tab:results} summarizes the performance recovery. 

% INSERT YOUR CSV DATA HERE MANUALLY OR USE THE FORMAT BELOW
\begin{table}[htbp]
\caption{Accuracy Recovery on CIFAR-10 (75\% Pruning)}
\begin{center}
\begin{tabular}{lcccc}
\toprule
\textbf{Model} & 
\parbox[b]{1.6cm}{\centering \textbf{Teacher} \\ \textbf{Acc. (\%)}} & 
\parbox[b]{1.6cm}{\centering \textbf{Pruned} \\ \textbf{Acc. (\%)}} & 
\parbox[b]{1.8cm}{\centering \textbf{Recovered} \\ \textbf{Acc. (\%)}} & 
\textbf{Improvement} \\
\midrule
ResNet18 & 93.28 & 73.29 & 93.10 & +19.81\\
ResNet34 & 93.68 & 76.03 & 93.51 & +17.48\\
ResNet50 & 93.05 & 82.33 & 92.07 & +9.74\\
\bottomrule
\end{tabular}
\label{tab:results}
\end{center}
\end{table}

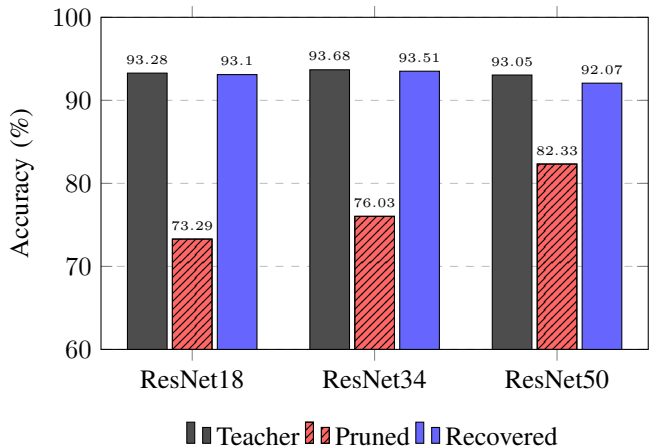
\begin{figure}[htbp]
\centering
\begin{tikzpicture}
    \begin{axis}[
        ybar,
        bar width=15pt,
        width=\columnwidth, % Fits the column width
        height=6cm,
        enlarge x limits=0.25,
        legend style={at={(0.5,-0.2)}, anchor=north, legend columns=-1, draw=none},
        ylabel={Accuracy (\%)},
        symbolic x coords={ResNet18, ResNet34, ResNet50},
        xtick=data,
        ymin=60, ymax=100, % Zoomed in to show the drop/recovery clearly
        nodes near coords,
        every node near coord/.append style={font=\tiny, rotate=0},
        ymajorgrids=true,
        grid style=dashed,
    ]

    % 1. Teacher Accuracy (Baseline - Gray/Black)
    \addplot[fill=black!70, draw=black] coordinates {
        (ResNet18, 93.28)
        (ResNet34, 93.68)
        (ResNet50, 93.05)
    };

    % 2. Pruned Accuracy (Drop - Red/Pattern)
    \addplot[fill=red!60, draw=black, postaction={pattern=north east lines}] coordinates {
        (ResNet18, 73.29)
        (ResNet34, 76.03)
        (ResNet50, 82.33)
    };

    % 3. Recovered Accuracy (Result - Blue)
    \addplot[fill=blue!60, draw=black] coordinates {
        (ResNet18, 93.10)
        (ResNet34, 93.51)
        (ResNet50, 92.07)
    };

    \legend{Teacher, Pruned, Recovered}
    \end{axis}
\end{tikzpicture}
\caption{Comparison of Accuracy before pruning, after 75\% pruning, and after Data-Free Recovery. The recovered models (Blue) nearly match the original Teacher performance (Gray).}
\label{fig:acc_chart}
\end{figure}

\subsection{Analysis}
The experimental results highlight several critical insights regarding the interplay between model capacity, sparsity, and data-free recovery.

\subsubsection{Robustness of Over-parameterized Models}
We observed a strong positive correlation between network depth and resistance to pruning. As shown in Table \ref{tab:results}, the deeper \textbf{ResNet50} retained the highest post-pruning accuracy (82.33\%) compared to the shallower \textbf{ResNet18} (73.29\%) at the same 75\% sparsity level. This aligns with the "Lottery Ticket Hypothesis," suggesting that deeper networks possess a higher degree of redundancy. Consequently, ResNet50 retains a more capable sub-network immediately after weight masking, requiring less aggressive recovery than its shallower counterparts.

\subsubsection{Efficacy of Synthetic Data}
Despite the lack of real training data, the recovery phase yielded substantial improvements across all architectures. \textbf{ResNet18} demonstrated the most significant gain, recovering by \textbf{+19.81\%} to reach a final accuracy of 93.10\%, effectively matching its unpruned teacher (93.28\%). This confirms that the synthetic "dream" images generated via DeepInversion successfully capture the essential feature statistics (mean and variance) required to retrain the student's decision boundaries.

\subsubsection{Recovery Saturation}
It is notable that while the pruning impact varied, the final recovered accuracy for all three models converged to within $\approx 1\%$ of their respective teachers. This suggests that the Knowledge Distillation process effectively saturates the capacity of the sparse sub-networks, restoring them to their maximum theoretical performance given the 75\% reduction in parameters. The method proves that accurate compression is achievable even in strict privacy-preserving environments where original data is unavailable.

\section{Discussion and Limitations}

\subsection{Computational Overhead}
The generation of synthetic data requires an optimization loop (backpropagation to the input) for every batch generated. In our experiments, generating 1,024 images required approximately 200 iterations per batch. While this is computationally more expensive than standard inference, it is a one-time cost. Once the synthetic dataset is generated, it can be reused for multiple distillation runs or different sparsity levels.

\subsection{Quality of Synthetic Data}
As seen in Figure \ref{fig:dreams}, DeepInversion images are not perfectly photorealistic. They often resemble "dream-like" textures that maximize the activation of specific filters. However, for the purpose of Knowledge Distillation, photorealism is not strictly necessary. The primary goal is to generate inputs that lie on the data manifold defined by the teacher's decision boundaries. Our results confirm that these texture-rich images are sufficient to transfer knowledge to the student.

\section{Conclusion and Future Work}
In this work, we demonstrated a fully data-free pipeline for recovering the accuracy of pruned neural networks. By synthesizing data via DeepInversion and applying Knowledge Distillation, we enable the compression of models in privacy-restricted environments. Future work can explore the generation of more diverse synthetic datasets using Generative Adversarial Networks (GANs) and the application of this method to structured pruning scenarios where entire channels are removed.

\section*{Acknowledgment}
This project was implemented as part of the coursework requirements. The code is open-sourced under the MIT License.

\end{document}